\documentclass[letterpaper, 10 pt, journal]{ieeeconf}

\IEEEoverridecommandlockouts

\usepackage{etextools}
\usepackage{amssymb}
\usepackage{float}
\usepackage{makeidx}
\usepackage{amsmath}
\usepackage{bbm}
\usepackage{epsfig}
\usepackage{epsf}
\usepackage{psfrag}
\usepackage{verbatim}
\usepackage{color}
\usepackage{multirow}
\usepackage{tabularx}
\usepackage{booktabs}
\usepackage[tight,footnotesize]{subfigure}
\usepackage{array}
\usepackage{soul}
\usepackage{footnote}
\usepackage{cite}
\usepackage{dblfloatfix}
\usepackage{tcolorbox}

\usepackage{color, colortbl}
\usepackage[colorlinks,bookmarksnumbered,citecolor=orange,urlcolor=orange]{hyperref}
\usepackage{graphicx}
\graphicspath{{Figures}}
\DeclareGraphicsExtensions{.pdf,.png}
\usepackage{bigstrut}
\usepackage[english]{babel}

\usepackage{csquotes}

\usepackage[T1]{fontenc}
\usepackage{algorithm, setspace}
\usepackage{algpseudocode}
\usepackage{url}
\usepackage{multirow}
\usepackage{float}
\usepackage{xcolor}
\usepackage{hyperref}
 \hypersetup{
     colorlinks=true,
     linkcolor=orange,
     filecolor=orange,
     citecolor=orange,      
     urlcolor=orange,
     }

\newcommand{\p}[1]{\smallskip \noindent \textbf{{#1}.}}
\newcommand{\eq}[1]{Equation~(\ref{eq:#1})}
\newcommand{\fig}[1]{Figure~\ref{fig:#1}}
\usepackage{balance}
\usepackage{amssymb}
\let\labelindent\relax
\usepackage{enumitem}

\title{\LARGE
RECON: Reducing Causal Confusion with Human-Placed Markers

}
\author{Robert Ramirez Sanchez$^{1}$, Heramb Nemlekar$^{1}$, Shahabedin Sagheb$^{1}$, Cara M.\ Nunez$^{2}$, and Dylan P.\ Losey$^{1}$\vspace{-1em}
\thanks{This work was supported by NSF Grants \#2337884 and \#2246446 \newline $^{1}$R.\ R.\ Sanchez, H.\ Nemlekar, S.\ Sagheb, and D.\ Losey are with the Collaborative Robotics Lab (\href{https://collab.me.vt.edu/}{Collab}), Dept. of Mechanical Engineering, Virginia Tech, Blacksburg, VA 24061. \newline $^{2}$C.\ M.\ Nunez is with the Sibley School of Mechanical and Aerospace Engineering, Cornell University, Ithaca, NY 14853. \newline Email: \texttt{robertjrs@vt.edu}}
}
\begin{document}

\maketitle

\begin{abstract}

Imitation learning enables robots to learn new tasks from human examples.
One fundamental limitation while learning from humans is \textit{causal confusion}.
Causal confusion occurs when the robot's observations include both task-relevant and extraneous information: for instance, a robot's camera might see not only the intended goal, but also clutter and changes in lighting within its environment.
Because the robot does not know which aspects of its observations are important \textit{a priori}, it often misinterprets the human's examples and fails to learn the desired task.
To address this issue, we highlight that --- while the robot learner may not know what to focus on --- the human teacher does.
In this paper we propose that the human proactively marks key parts of their task with small, lightweight \textit{beacons}.
Under our framework (RECON) the human attaches these beacons to task-relevant objects before providing demonstrations: as the human shows examples of the task, beacons track the position of marked objects.
We then harness this offline beacon data to train a task-relevant state embedding.
Specifically, we embed the robot's observations to a latent state that is correlated with the measured beacon readings: in practice, this causes the robot to autonomously filter out extraneous observations and make decisions based on features learned from the beacon data.
Our simulations and a real robot experiment suggest that this framework for human-placed beacons mitigates causal confusion.
Indeed, we find that using RECON significantly reduces the number of demonstrations needed to convey the task, lowering the overall time required for human teaching.
See videos here: \url{https://youtu.be/oy85xJvtLSU}

\end{abstract}

\section{Introduction} \label{sec:intro}

\begin{figure*}[t]
	\begin{center}
 		\includegraphics[width=0.9\linewidth]{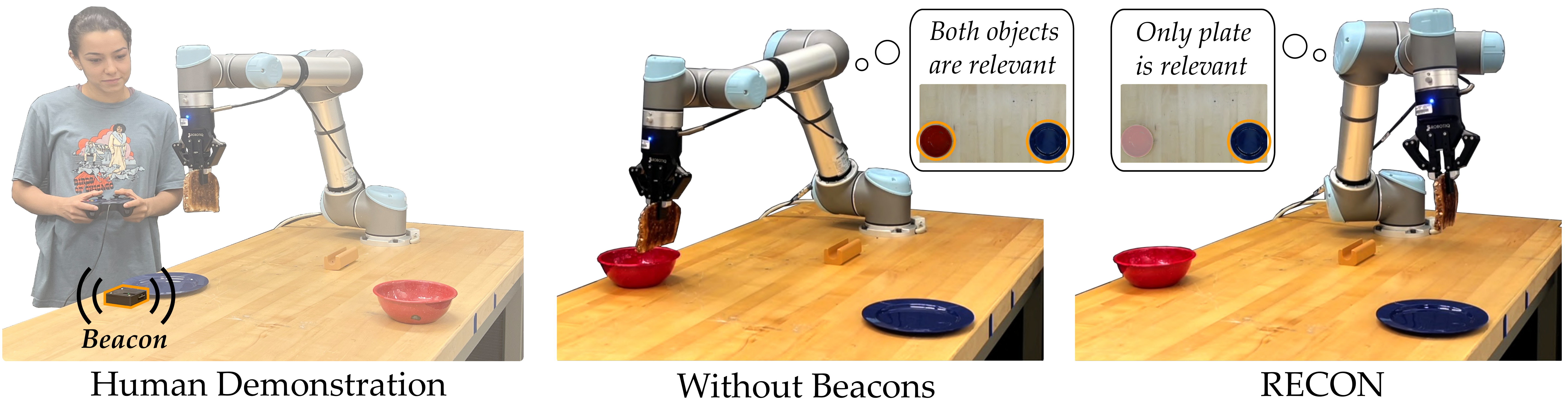}
        \vspace{-2ex}
		\caption{(Left) Human teaching a robot arm to place bread on a blue plate. The environment also includes a red bowl that is not relevant to the task. The robot observes this scene from a top-down camera. We enable humans to mark task-relevant objects (i.e., the plate) with \textit{beacons} --- lightweight devices that track the position of the marked objects while the human is demonstrating the task. (Center) A robot trained to imitate the task with only visual information incorrectly infers that both objects are relevant and delivers the bread to the wrong location. (Right) A robot trained with beacons using our RECON algorithm is able to mitigate this confusion and learns to correctly drop the bread onto the plate, even after the beacons are removed.}
		\label{fig:front}
	\end{center}
	\vspace{-1.5em}
\end{figure*}

Imagine a human teaching their personal robot arm to pick up bread and drop it on a plate (see \fig{front}).
To learn this new task, the robot observes its environment and records the human's demonstrations.
But not all the information it observes is relevant: in addition to the bread and plate, the robot's camera also sees a bowl, assorted objects in the background, and even changes in the lighting conditions.
These high-dimensional and complex real-world observations make it challenging for the robot to learn the desired task.
For instance, if the position of the bowl changes, should the robot's actions change? 
What about the objects in the background or the scene's lighting --- should these features affect the robot's behavior?
Without additional information, the robot does not know \textit{a priori} which aspects of its observations contain task-critical information.
Unintentionally focusing on the wrong thing can result in incorrect learning; e.g., moving the bread to a bowl in the background instead of the plate.
More generally, imitation learning with real-world observations can produce \textit{causal confusion}~\cite{de2019causal} and \textit{compounding errors}~\cite{spencer2021feedback}, leading to robots that fail to learn the demonstrated task.

Existing research often tries to mitigate these issues by developing purely algorithmic solutions.
Current methods leverage attention networks, autoencoders, and object-centric priors to try and autonomously extract a task-relevant state from the robot's high-dimensional observations~\cite{osa2018algorithmic, zhu2023viola, gao2023k, jonnavittula2024view, vaswani2017attention, pari2021surprising, ravichandar2020recent}.
For example, in \fig{front} the state learned by these approaches could include the pixel location of all objects that the robot detects.
While today's algorithmic methods can learn compact representations of the robot's observations, this process requires data~\cite{lesort2018state}: the human needs to spend more time giving additional examples so that the robot can indirectly infer what parts of the environment are relevant for the task. 

In this paper we propose to mitigate causal confusion and efficiently learn task-relevant representations by 
directly leveraging the human teacher's knowledge of the key task items.
Specifically, we enable the human to add \textit{beacons} (Bluetooth tracking devices) to the environment before giving demonstrations.
Beacons are small, lightweight, and inexpensive.
As the human demonstrates the task, these beacons automatically record the position of their marked items (e.g., the bread and the plate).
Our intuition is that:
\begin{center}\vspace{-0.3em}
\textit{Humans will attach beacons to task-relevant objects}.
\vspace{-0.3em}
\end{center}
Put another way, we hypothesize that the beacons are correlated with aspects of the observation the robot should focus on.
We apply our insight to develop \textbf{RECON}: \textbf{Re}ducing Causal \textbf{Con}fusion with Human-Placed Markers, an imitation learning algorithm that (a) learns a state representation to embed the beacon data, and then (b) learns a policy that maps this compact state representation to robot actions.
Returning to our kitchen example: when the human teacher adds beacons to the bread and plate, the robot learns to extract features involving the bread and plate from its camera images, and then take actions based on those features.

Overall, we make the following contributions:

\p{Leveraging Beacons}
We extend imitation learning from offline human demonstrations to include user-placed beacons.
Under this formalism, beacons provide an added source of supervision for extracting the task-relevant states (i.e., the features) from robot observations.

\p{RECON}
We introduce our RECON algorithm for learning a robot policy from human demonstrations.
The robot (a) minimizes its divergence from the human's demonstrated actions and (b) maximizes mutual information between the features and beacon readings.
One key advantage of our algorithm is that beacons are not required after training.

\p{Experiments}
We conduct multiple simulations and a robot arm experiment
in visual imitation learning settings.
When the human places the beacon(s) on relevant objects, we find that our RECON algorithm learns the desired task accurately in fewer demonstrations than previous imitation learning approaches.
While deploying beacons requires added time, it is considerably less than providing more demonstrations, and thus RECON reduces the total training time. 
\section{Related Work} \label{sec:related}

Our research explores imitation learning settings where the human teacher adds markers to the environment to measure the key aspects of their task demonstrations.

\p{Imitation Learning} 
Within imitation learning, the robot uses examples from a human expert to extrapolate the desired mapping between observations and actions~\cite{osa2018algorithmic}.
When the observations are high-dimensional (e.g., images), it is often challenging for the robot to determine which aspects of those observations should affect its actions~\cite{de2019causal}.
Accordingly, recent works that learn from human demonstrations have tried to extract succinct, task-critical representations of visual observations.
For instance, object-centric approaches autonomously detect items in the scene (e.g., the bread, the toaster, the bowl), and then condition the robot's policy on those objects~\cite{zhu2023viola, gao2023k, jonnavittula2024view}.
More generally, robots can leverage attention networks~\cite{vaswani2017attention} and representation learning~\cite{pari2021surprising} to learn the key features of an image given multiple demonstrations.
Overall, these methods attempt to accelerate the robot's learning by \textit{indirectly} estimating task-critical data from high-dimensional observations.
But the robot could learn more efficiently if it had \textit{direct} access to this information (e.g., if the robot knew the bread and plate positions).

\p{Additional Sensors}
To more directly measure the relevant components of the state, other imitation learning research explores additional sensors and interfaces~\cite{ravichandar2020recent}.
Consider our motivating example of putting bread onto a plate; by tracking where the human looks when teaching the system (e.g., by detecting that the human gazes at the bread and then the plate), the robot can determine which features of the environment it should focus on.
More generally, related works have developed multiple types of sensors and interfaces: these include gaze tracking~\cite{biswas2024gaze, saran2021efficiently}, augmented reality~\cite{jiang2024comprehensive, quintero2018robot, luebbers2021arc},~crowdsourced labels \cite{allevato2020learning}, and manually specified waypoints~\cite{akgun2012keyframe, belkhale2023hydra}.
Most relevant to our proposed approach are physical sensors that the human uses or wears when interacting with the environment.
In~\cite{young2021visual, song2020grasping} the human utilizes an instrumented grabber to manipulate objects, in~\cite{wei2024wearable} the human wears an articulated glove while providing demonstrations, and in~\cite{song2023data} the human moves around a position and optical tracker to trace their desired trajectory.
These existing sensors focus on the human, and often must be present during both training and task execution.
By contrast, we propose to instrument the \textit{environment} only when collecting demonstrations, and then \textit{remove} those markers when the robot acts autonomously.
\section{Problem Statement} \label{sec:problem}

We consider contexts where a robot arm is learning a new task from human demonstrations.
The robot observes its environment (e.g., using a camera), and the human teacher kinesthetically guides or teleoperates the robot arm through instances of the desired task.
The robot learner does not know \textit{a priori} which aspects of the environment it should focus on when learning the task.
To accelerate the robot's learning, we enable the human to instrument the environment by adding beacons to mark key items.
Overall, the robot's objective is to leverage these beacons to efficiently learn a policy that maps its environment observations to actions.
Below we formalize each aspect of this problem:

\p{Robot}
The robotic system consists of a robot arm and RGB camera.
The state of the robot arm is $x \in \mathbb{R}^n$, and the robot takes actions $u \in \mathbb{R}^n$.
For instance, within our motivating example, $x$ is the robot's joint position, and $u$ is the robot's joint velocity.
We emphasize that $x$ only captures information about the robot arm; in order to observe the state of the environment, the robot uses its RGB camera.
Let $y \in \mathbb{R}^m$ be the current camera reading (e.g., an image of the toast, bowl, and plate).
Overall, at every timestep the robot observes a high-dimensional vector $(x, y)$ and takes action $u$.

\p{Features}
The observed state $(x, y)$ contains the relevant information for learning the task.
But this high-dimensional state may also include unnecessary data that should not affect the robot's behavior.
Consider our kitchen example: the camera sees the bowl in image $y$, but the position of that bowl should have no impact on how the robot picks up and moves the bread.
More formally, let $\phi = f(x, y)$ be a minimal representation of the observed state that is sufficient for the desired task.
We instantiate $\phi \in \mathbb{R}^k$ as a feature vector of length $k$, where this feature dimension $k$ is less than the image dimension $m$.
Returning to our example, $\phi$ could contain the location of the toast and the location of the plate.
The robot does not have access to feature vector $\phi$; the robot observes $x$ and $y$, but it does not know \textit{a priori} which aspects of these observations are features of the desired task.

\p{Beacons}
We enable the human teacher to intuitively show the robot which aspects of its observations are important by introducing physical beacons.
These beacons are markers: the human can attach them to objects in the environment, and each beacon will continuously relay its real-time position and distance relative to the other beacons.
In our experiments, we use beacons with the same technology as Apple AirTags~\cite{airtag}. Specifically, we use Qorvo's DWM1001-DEV ultra-wideband (UWB) transceiver development boards~\cite{decawave}.
These beacons utilize a combination of Bluetooth and UWB to measure position despite of visual occlusions (i.e., the robot does not need to see the beacons).
More generally, our algorithmic framework is \textit{not tied to any specific type of beacon}.
We simply assume that the beacons provide measurements of the form $b = g(x, y)$, where $b \in \mathbb{R}^d$ consists of information relayed from one or multiple beacons.
Vector $b$ depends on the robot state $x$, the observed environment $y$, and the locations where the human places the beacon(s).

\p{Policy}
The human marks the key objects in the environment at the start of the teaching process and then demonstrates the desired behavior.
During these demonstrations the robot records a dataset of observed states, beacon measurements, and expert actions: $\mathcal{D} = \{(x, y, b, u)\}$.
After the demonstrations are complete, the human removes the beacons.
The robot's objective is to learn to autonomously perform the demonstrated task.
Specifically, the robot should learn a policy $u = \pi(x, y)$ that maps observations to actions.
We note that this policy does not depend on beacon readings $b$ --- 
the robot only uses beacons to scaffold its learning.
\section{Leveraging Beacons to Supervise Learning} \label{sec:method}

In this section we present our approach for using demonstrations with human-placed beacons to efficiently bootstrap robot policies.
We build upon our underlying hypothesis: beacon readings are correlated with the parts of the observation the robot should focus on.
Utilizing this insight, in Section~\ref{sec:M1} we first outline a model structure that connects observations, features, beacons, and actions.
Next, in Section~\ref{sec:M2} we derive loss functions that cause the learned policy to match the human's demonstrations (minimizing divergence), while also correlating the learned features with the measured beacons (maximizing mutual information).
In Section~\ref{sec:M3} we finally combine these loss functions to reach RECON, an imitation learning algorithm that reduces causal confusion by biasing the robot's state representation towards task-relevant features.
Once trained, \textit{the robot's policy 
can be used without any beacons in the environment}.

\begin{figure}[t]
	\begin{center}
 		\includegraphics[width=0.95\linewidth]{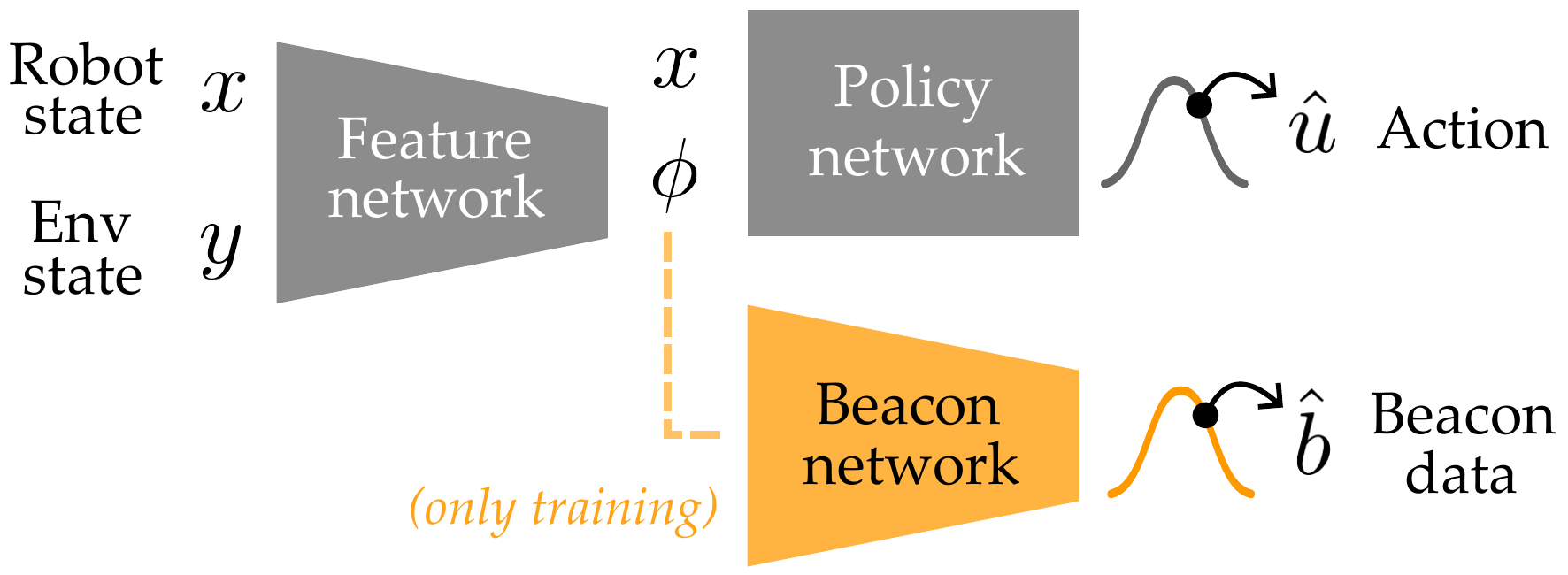}
        \vspace{-0.5em}
		\caption{Our proposed model architecture consists of a \textit{feature network} that maps observations $(x, y)$ to task-relevant features $\phi$, a \textit{beacon network} that relates the features to beacon data $b$, and a \textit{policy network} that estimates actions $u$ based on the robot state and learned features. The beacon network is only utilized at training time to supervise the task-relevant features.}
		\label{fig:method}
	\end{center}
	\vspace{-1.5em}
\end{figure}

\subsection{Model Structure} \label{sec:M1}

An outline of our proposed model structure is shown in \fig{method}. Below we explain the three component networks:

\p{Policy Network} 
At training time the robot has access to the offline dataset of task demonstrations $\mathcal{D} = \{(x, y, b, u)\}$.
Given this dataset, the robot's objective is to learn a policy that imitates the human expert and maps observations $(x, y)$ to actions $u$.
We recognize that not all of the information in the observed state $(x, y)$ is relevant for the desired task; to correctly imitate the human, the robot only needs to reason over the task-relevant features $\phi = f(x, y)$.
Accordingly, we propose to learn a \textit{policy network} of the form: $\pi_{\theta_1}(u \mid x, \phi)$.
This policy is a network with weights $\theta_1$. 
In practice, $\pi$ determines the probability of action $u$ given robot's current state $x$ and the estimated features $\phi$.

\p{Feature Network}
The robot learner does not know the features of the task \textit{a priori}.
Instead, the robot must learn to extract these features $\phi$ from its state observations $(x, y)$.
Consistent with our definitions from Section~\ref{sec:problem}, we therefore introduce a \textit{feature network} $\phi = f_{\theta_2}(x, y)$ with weights $\theta_2$.
This network inputs observations $x$ and $y$ and deterministically outputs a feature vector $\phi$.

\p{Beacon Network}
The final component of our model structure connects the features and beacons.
Here we apply our insight that the beacon data $b$ should be correlated with the unknown features $\phi$.
In order to correlate $b = g(x, y)$ and $\phi = f(x, y)$, we add a \textit{beacon network} $h_{\theta_3}(b \mid \phi)$.
This network with weights $\theta_3$ outputs the probability of a beacon measurement $b \in \mathbb{R}^d$ given that the true features are $\phi \in \mathbb{R}^k$.
Here the choice of dimension $k$ is important.
Although we assume that $b$ is correlated with the task-relevant features, we recognize that the beacon readings may not capture all aspects of the unknown $\phi$.
Hence, we constrain the designer to select $k > d$, so that the learned vector $\phi$ can contain additional information beyond the beacon readings $b$.

\subsection{Loss Functions} \label{sec:M2}

We use two sources of supervision to train our model: the expert actions and beacon readings.
Similar to behavior cloning, we update the policy and feature networks so that the robot's learned actions match the actions provided by the human expert~\cite{spencer2022expert}.
But we also apply the beacon readings to directly supervise feature extraction; because we hypothesize that $b$ and $\phi$ should be correlated, we can leverage $b$ to guide the robot towards $\phi$.
Below we derive the loss functions that result from both types of supervision:

\p{Action Supervision}
Let $\pi^*(u \mid x, y)$ be the ideal, unknown policy that the human teacher wants the robot to follow.
The robot should learn to match this policy.
More concretely, we seek to minimize the Kullback–Leibler (KL) divergence between $\pi^*$ and $\pi$ across the demonstration dataset:
\begin{equation} \label{eq:M1}
    D_{KL}(\pi^* \mid\mid \pi) = -\mathbb{E}_{(x, y, u) \sim \mathcal{D}} \big[\log \pi_{\theta_1}(u \mid x, \phi)\big] + C_1
\end{equation}

Here $(x, y, u) \in \mathcal{D}$ are observation-action samples from the ideal policy $\pi^*$ that the human provides during their demonstrations.
The constant $C_1$ in \eq{M1} is the entropy of $\pi^*$, and this term does not depend on $\pi$.
Hence, to minimize the KL divergence and match the ideal policy, our model should learn to minimize the following loss function:
\begin{equation} \label{eq:M2}
    \mathcal{L}_1(\theta_1, \theta_2) = -\mathbb{E}_{(x, y, b, u) \sim \mathcal{D}} \Big[\log {\pi_{\theta_1}\big(u \mid x, f_{\theta_2}(x, y)\big)}\Big]
\end{equation}
When moving from \eq{M1} to \eq{M2} we have substituted in our feature network $\phi = f_{\theta_2}(x, y)$.
We have also added a marginalization over $b$  --- without affecting the loss --- because this term is not used within the expectation.

Intuitively, minimizing \eq{M2} means that the system has learned a policy $\pi$ and feature encoding $f$ that causes the robot to mimic the human teacher's actions across dataset $\mathcal{D}$.
But we do not know how this robot will perform when it encounters new states at run time~\cite{mehta2024stable}.

\p{Beacon Supervision}
To facilitate learning --- and better infer what aspects of the state observations the robot should focus on --- we introduce a second loss function.
This loss is based on the correlation between beacon readings $b$ and learned features $\phi$.
Specifically, we seek to maximize the mutual information between $b$ and $\phi$ across dataset $\mathcal{D}$:
\begin{equation} \label{eq:M3}
    I(b \, ; \, \phi) = \mathbb{E}_{(x, y, b) \sim \mathcal{D}} \Bigg[\log{\frac{P(b \mid \phi)}{P(b)}}\Bigg]
\end{equation}
Increasing mutual information $I(b ; \phi)$ means that, if we observe variable $b$, we gain information about $\phi$ (and vice versa).
We can rewrite this mutual information in terms of our networks.
Substituting in the beacon network $P(b \mid \phi) = h_{\theta_3}(b \mid \phi)$ and the feature network $\phi = f_{\theta_2}(x, y)$, we reach:
\begin{equation} \label{eq:M4}
    I(b \, ; \, \phi) = \mathbb{E}_{(x, y, b) \sim \mathcal{D}} \Big[\log {h_{\theta_3}\big(b \mid f_{\theta_2}(x, y)\big)}\Big] + C_2
\end{equation}
Similar to $C_1$ in \eq{M1}, here constant $C_2$ does not depend on our model and can be dropped from the analysis.
By remembering to flip the sign (because we want to maximize mutual information gain but minimize loss) and then adding in a marginalization over action $u$, we obtain our second loss function for supervising the features:
\begin{equation} \label{eq:M5}
    \mathcal{L}_2(\theta_2, \theta_3) = -\mathbb{E}_{(x, y, b, u) \sim \mathcal{D}} \Big[\log {h_{\theta_3}\big(b \mid f_{\theta_2}(x, y)\big)}\Big]
\end{equation}

In practice, minimizing \eq{M5} causes the robot to learn a feature vector $\phi$ that can be decoded to recover the beacon signal $b$.
This does not mean that $\phi$ is equal to $b$ --- instead, we obtain a function $h$ that maps from $\phi$ to $b$.

\subsection{RECON Algorithm} \label{sec:M3}

Now that we have defined the model structure (Section~\ref{sec:M1}) and loss functions (Section~\ref{sec:M2}), we are ready to introduce our RECON algorithm.
The human teacher first instruments the environment by adding beacons, and then the human provides task demonstrations $\mathcal{D}$.
The robot trains its policy network $\pi_{\theta_1}$, feature network $f_{\theta_2}$, and beacon network $h_{\theta_3}$ to minimize the combined loss across the dataset:
\begin{multline} \label{eq:M6}
    \mathcal{L}(\theta_1, \theta_2, \theta_3) = -\mathbb{E}_{(x, y, b, u) \sim \mathcal{D}} \Big[\log {\pi_{\theta_1}\big(u \mid x, f_{\theta_2}(x, y)\big)} \\ + \log {h_{\theta_3}\big(b \mid f_{\theta_2}(x, y)\big)}\Big]
\end{multline}
where \eq{M6} is the sum of the action supervision in \eq{M2} and the beacon supervision in \eq{M5}.
All three networks that compose our model are trained simultaneously using this loss function.
Minimizing \eq{M6} trains the model so that (a) we can predict the beacon reading from our learned features $\phi$, and (b) we can reconstruct expert actions given $x$ and $\phi$.
As compared to state-of-the-art imitation learning baselines, the key novelty of RECON is the additional supervision provided by the beacon markers.
This supervision guides the robot's feature extraction such that the learned features are correlated with the human's marked measurements, resulting in a state representation $\phi$ that is aligned with the desired task.
Returning to our kitchen example: because $\phi$ is trained to capture the position of the bread and plate from environment images $y$, the robot learns to make decisions based on these two objects, and ignores the color and location of the extraneous bowl.


\p{Implementation} 
In our experiments, the feature network $f$ was a convolutional neural network (CNN) that mapped from image $y$ to vector $\phi$.
Both the beacon network $h$ and the policy $\pi$ were multilayer perceptrons (MLPs) with fully connected linear layers and rectified linear unit (ReLU) activation functions. 
The networks were trained using the Adam optimizer at a learning rate of $0.0001$.
Our code can be found here:
\url{https://github.com/VT-Collab/RECON}

We emphasize that \textit{our proposed loss in \eq{M6} is not constrained to CNNs or MLPs}, and designers can replace them with other architectures such as transformers.

\begin{figure*}[ht]
    \begin{center}
    \includegraphics[width=0.95\textwidth]{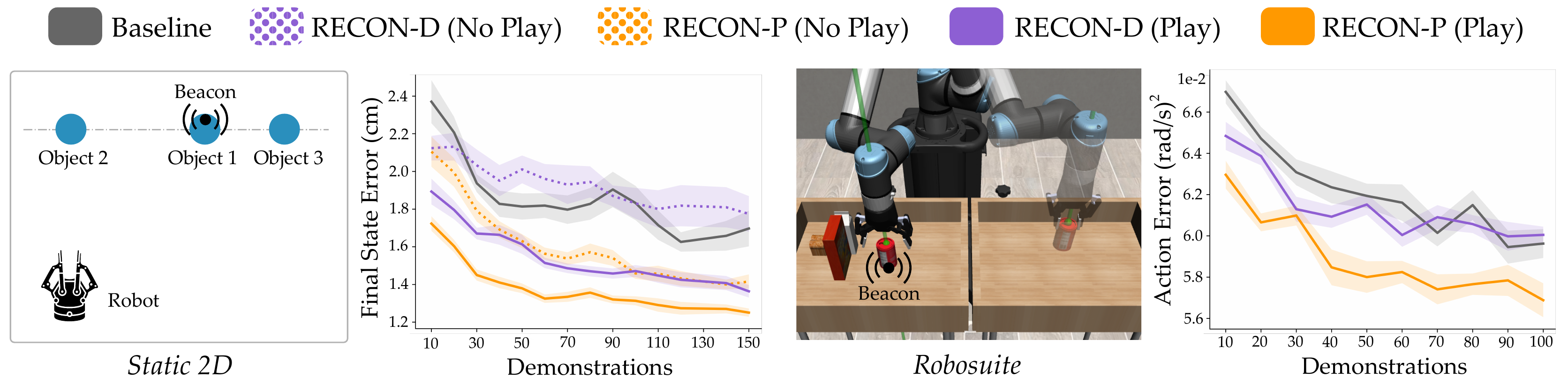}
    \vspace{-0.75em}
    \caption{Environments and results for our simulations in Section~\ref{sec:sim1} averaged over $20$ training and testing runs. (Left) In the \textit{Static 2D} environment, the robot must reach the object in the center. Here a robot trained with position beacons (RECON-P) reaches closer to the target than a robot trained without beacons (Baseline). However, when using distance beacons (RECON-D) we require additional Play data (i.e., state and beacon pairs) to properly calibrate the feature network and ensure that the robot performs better than the baseline. (Right) In the \textit{Robosuite} environment, the robot must transfer the right-most object from one bin to the other. Here the robot imitates the expert policy more closely when trained with position beacons and play data (RECON-P (Play)), while training with distance beacons and play data (RECON-D (Play)) produces results similar to the Baseline.}
    \vspace{-1.5em}
    \label{fig:state}
    \end{center}
\end{figure*}

\section{Simulation Experiments}\label{sec:sim}

We start by evaluating our proposed algorithm in simulated tasks. Here we test whether robots can learn the desired tasks more accurately from expert demonstrations by leveraging task-relevant beacon information.



\begin{figure*}[ht]
    \begin{center}
    \includegraphics[width=0.9\textwidth]{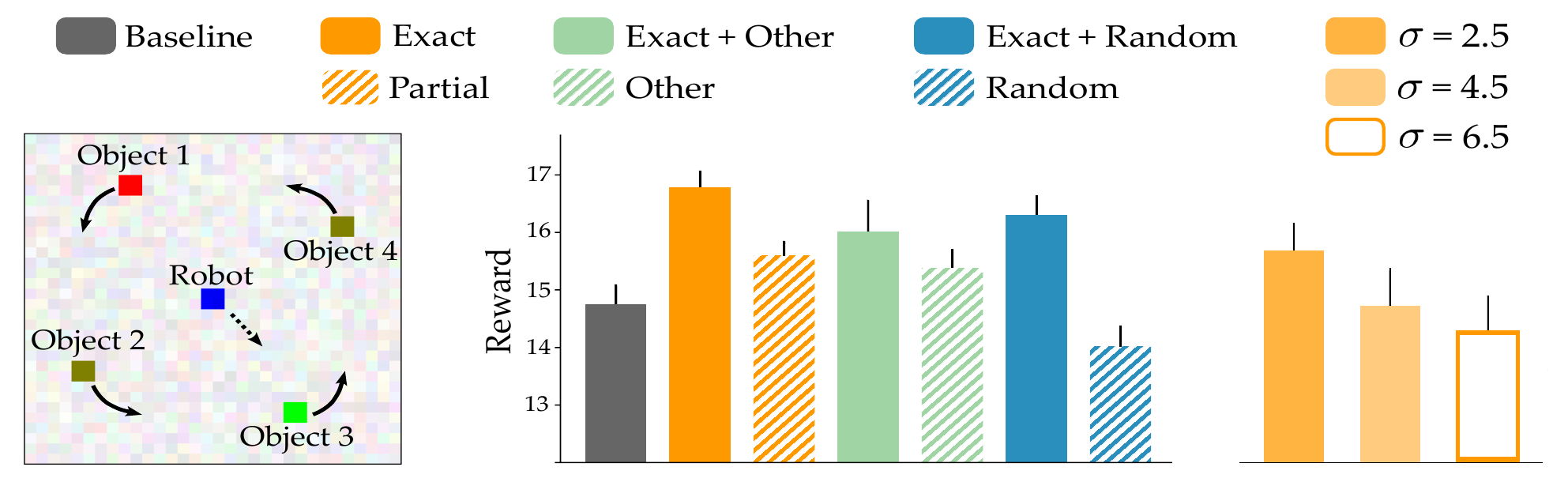}
    \vspace{-0.75em}
    \caption{Simulation results averaged over $15$ training and testing runs in the \textit{Dynamic 2D} environment. We find that robots trained with beacons attached to the task-relevant objects (Exact) achieve the highest rewards while random beacon placement results in the lowest. Though Partial and Other (less relevant) beacon readings also perform worse than Exact, they still outperform the Baseline (which does not use beacons). Exact remains better than the Baseline even with moderate noise ($\sigma=2.5$), only performing worse at high noise levels ($\sigma > 4.5$).}
    \vspace{-1.5em}
    \label{fig:dynamic}
    \end{center}
\end{figure*}

\p{Environments} We evaluate our algorithm in three environments: \textit{Static 2D}, \textit{Robosuite}, and \textit{Dynamic 2D}. The \textit{Static 2D} environment consists of three objects placed on a line as shown in \fig{state}-Left. In each instantiation of the task, the objects are shuffled and positioned randomly. The robot's goal is to reach and grasp the object in the center. The \textit{Robosuite} environment has two bins. The left bin contains four objects that are randomly positioned at the start of the task, as shown in \fig{state}-Right. The robot's goal is to pick up the right-most object and transfer it to the other bin. In both \textit{Static 2D} and \textit{Robosuite}, the objects are static until grasped by the robot. In contrast, the objects in the \textit{Dynamic 2D} environment move anticlockwise in a circle (see \fig{dynamic}). The robot's goal is to move away from the red object while disregarding the other objects.

\subsection{Imitation Learning with Task-Relevant Beacons}\label{sec:sim1}

First we assess whether using beacons to mark the relevant objects during training can enable the robot to mitigate causal confusion and learn a better policy from task demonstrations. In these experiments we test our algorithm in the \textit{Static 2D} and \textit{Robosuite} environments, and assume that beacons are correctly placed on the relevant objects, i.e., the central object in \textit{Static 2D} and the right-most object in \textit{Robosuite}.

\p{Demonstrations} 
We collect demonstrations by simulating the robot actions until task completion in both environments. At each time step in \textit{Static 2D}, we record the robot's position $x$, the positions of the three objects $y$, the beacon reading $b$, and a unit action $u$ in the direction of the central object. Demonstrations in \textit{Static 2D} have at most 10 time steps. In \textit{Robosuite}, the actions are generated by an expert policy. At each time step, we record the robot's 6D joint angles $x$, the position of all objects in the bin $y$, the beacon reading $b$, and joint velocity $u$.
After each demonstration, the environments are reset with the objects in random positions.


\p{Play data}
Since the objects in our environments are static, the beacon readings stay constant until the robot moves the target object. As a result, the beacon data recorded with the demonstrations can be insufficient for training the feature and beacon networks. To supplement the demonstration data, we collect an additional dataset of $(x, y, b)$ pairs by randomly initializing the environment and recording beacon readings for the task-relevant objects. For example, in \textit{Robosuite}, we randomize the object positions $y$ and measure the position of the right-most object $b$. We use this play data to further train the feature network with the loss in \eq{M5}.

Note that we gather play data without simulating the robot, i.e., we do not acquire extra demonstrations. In practice, play data is collected once to learn the mapping $f_{\theta_{2}}(x, y) = \phi$ and used to teach multiple tasks.

\p{Independent variables} As a baseline, we compare RECON to a standard imitation learning approach with feature and policy networks (the grey blocks in \fig{method}). The \textbf{Baseline} does not use any beacon data, instead, the feature network is trained end-to-end with the policy network by minimizing the loss \eq{M2}~\cite{finn2017one} to learn an unsupervised representation of the environment states. 

Our beacons transmit two types of data: distance and position. While distance requires less instrumentation, position is more informative. We separately evaluate our algorithm with both data types: \textbf{RECON-D}istance and \textbf{RECON-P}osition. We also evaluate training with and without play data: \textbf{RECON (Play)} and \textbf{RECON (No Play)}.


\p{Results}
\fig{state} summarizes our results. In \textit{Static 2D}, we test the learned policies in $100$ random task configurations by rolling out the actions for $10$ time steps and measuring the final distance between the robot and the target object. In \textit{Robosuite}, we measure the mean squared error between the actions of the expert policy and the actions of the learned policies in $110$ task configurations. The results in both environments are averaged over 20 training runs.

In \textit{Static 2D}, RECON-P achieves a lower final state error than the Baseline, with and without training on the play data. Meanwhile, RECON-D (No Play) performs similar to the Baseline and requires play data to perform better. 
RECON-P performs well even without play data because the mapping from the environment state to beacon position, $y \mapsto \phi \mapsto b$, is straightforward --- since $y$ already includes the position of the tagged object. As a result, $\phi$ can easily learn to recover $b$ from just the demonstration data. On the other hand, the mapping from $y$ to distance is indirect and requires additional play data to train $\phi$ effectively.

In \textit{Robosuite}, only RECON-P (Play) outperforms the Baseline, while all other approaches have similar action error. Unlike \textit{Static 2D}, where the objects are aligned in a straight line, the objects in this environment are scattered within a bin. Hence, the robot's distance to an object does not always correlate with its position, which explains why the performance does not improve when using beacons that transmit distance instead of position. 

Overall, these results suggest that when beacon data aligns perfectly with the task-relevant features, our algorithm learns better policies with the same amount of demonstrations compared to learning solely from the environment observations.

\subsection{Learning from Imperfect Beacon Data}\label{sec:sim2}

In our previous simulations we assumed that the beacons were accurately placed on the task-relevant objects. However, in real-world settings, users can make mistakes, e.g., placing beacons on irrelevant items. Moreover, these beacon readings can be noisy in practice. We now evaluate how our algorithm performs with imperfect beacon data.

\p{Independent variables}
We compare the same \textbf{Baseline} to learning from beacons placed on all relevant items (\textbf{Exact}), half of the relevant items (\textbf{Partial}), other less relevant items (\textbf{Other}), and random items (\textbf{Random}). For instance, in \textbf{Exact} placement the beacon transmits the $X$-$Y$ position of the task-relevant object (e.g., the red object), while in \textbf{Partial}, the beacon only transmits its position along the $X$ or $Y$ axis. In \textbf{Other}, the beacon is placed on an object adjacent to the red object and transmits its 2D position. Lastly, in \textbf{Random}, the beacon is placed on a randomly chosen object before each demonstration. We also compare with combinations of the above: \textbf{Exact+Other} and \textbf{Exact+Random}.
In addition to evaluating our approach with alternative beacon placements, we test the performance of our algorithm with noisy beacons: we add a zero mean Gaussian noise with standard deviations $\sigma = [2.5, 4.5, 6.5]$ to the \textbf{Exact} beacon readings.

\p{Demonstrations}
We collect demonstrations in the Dynamic 2D environment by randomly initializing the objects and simulating the robot's actions for 10 time steps. As opposed to the static environments where the observations were vectors of object positions, the observations in this environment are \textit{RGB images}. Because the objects in this environment move at each time step, we obtain diverse beacon readings in the demonstration data itself. Hence, we do not require any play data for training the feature and beacon networks.

\p{Results}
Our results are summarized in \fig{dynamic}. We train robot policies using RECON for each beacon placement and noise level. All policies are trained with $10$ demonstrations. We then test the learned policies in $100$ random task configurations by rolling out the actions for $10$ time steps and measuring the final distance between the robot and the red object. Since the robot's goal is to move away from the red object, the final distance corresponds to its \textbf{Reward}.

As expected, the rewards decrease with increasing noise in the beacon readings. The robot achieves the highest rewards when its policy is trained with perfect beacon data, i.e., exact placement with zero noise. The performance reduces as we use less relevant or noisy beacon data, with the lowest rewards for policies trained using randomly placed beacons. 
Overall, all beacon placements except Random result in higher rewards than the Baseline, which does not use any beacons.
These results indicate that the robot's performance will improve as long as there is some correlation between the beacon readings and the task-relevant features.


\section{Robot Experiment} \label{sec:robot}

In this section we evaluate our algorithm in a real-world environment. Specifically, we train a UR5 robot arm to pick up a piece of toast and drop it on a plate. We compare our RECON approach to the same Baseline as in our simulations. Unlike our simulated experiments, real-world settings require humans to spend additional time attaching beacons to task-relevant objects. 
Here we test if deploying physical beacons reduces the overall time required to train the robot.

\begin{figure}[t]
	\begin{center}
    \vspace{-0.75em}    
    \includegraphics[width=0.85\linewidth]
    {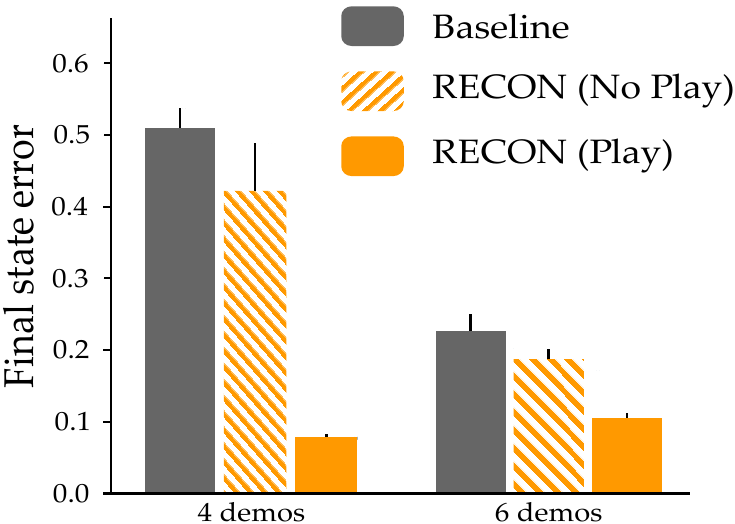}
    \caption{Results for our robot experiment averaged over $3$ end-to-end runs. With just four demonstrations, a robot trained without beacons (Baseline) fails to recognize the task-relevant objects (blue plate), resulting in poor performance during testing. In contrast, a robot trained with beacon readings (RECON (Play)) learns to focus on the plate and deliver the toast accurately.}
    \label{fig:robot_results}
	\end{center}
	\vspace{-1.5em}
\end{figure}



\p{Experimental setup}
The environment includes a blue plate, a red bowl, and a piece of toast placed on a table as shown in \fig{front}. At test time, the blue plate can lie in two locations: on the left or right edge of the table, while the red bowl is placed anywhere at random. An expert human teacher places a beacon near the blue plate, and then teleoperates the robot from the toast to the blue plate using a joystick. During each demonstration we record the robot joint states $x$, joint velocities $u$, images from a top-down camera $y$, and position readings from the beacon near the plate $b$.

\p{Demonstrations} We collect $2$ examples each for $3$ different plate and bowl configurations --- $6$ in total. Each demonstration contains ${\sim}150$ $(x, y, b, u)$ samples recorded at 10 Hz. 
On average it took $16.6$ seconds for a human expert to provide one demonstration and $2.3$ seconds to attach a beacon to the plate. 
We also collect $4$ seconds of play data by attaching the beacon to the plate and moving the bowl around. The play data contains $40$ $(y,b)$ pairs recorded at $10$ Hz.

\p{Results}
Our results are displayed in \fig{robot_results}. We train the robot policy with the collected demonstrations and test by executing the learned policy in the real world for $20$ different plate and bowl positions.

After training with only $4$ demonstrations, the Baseline learns to associate the robot's actions with the positions of both objects. As a result, it fails to generalize when the position of the bowl changes during testing. By contrast, RECON learns to ignore the bowl and focus on the position of the plate by leveraging the beacon readings (especially after training with play data). With more demonstrations, the Baseline gradually learns that the bowl is irrelevant to the task, bringing its performance closer to that of RECON (Play). Overall, these results are consistent with the findings from our simulations and highlight how human-placed beacons can mitigate causal confusion in imitation learning.

\p{Data collection time} 
To provide $4$ demonstrations, a human working with the Baseline will spend ${\sim}66.4$ seconds interacting with the robot, while with RECON (Play), the human will spend ${\sim}75$ seconds. The extra time includes $4$ seconds of play data and $4.6$ seconds to place the beacon in each of the two positions of the plate. 
However, without using beacons, the robot does not learn to place the toast accurately after only $4$ demonstrations. The Baseline requires more than $6$ demonstrations to perform similar to RECON (Play), which increases its training time by more than $33.2$ seconds. This shows how \textit{RECON reduces the total time required to train robots, even with the added step of attaching beacons}.

\section{Conclusion}

In this paper we equipped an expert human teacher with position tracking beacons.
The human attached these beacons to relevant objects, and we developed an imitation learning approach that synthesized both the human's demonstrations and the beacon data to learn the desired task.
Specifically, we trained the robot to encode its observations (e.g., camera images) into a compact state representation that maximized mutual information with the beacon data.
Our simulations and experiment suggest that this approach mitigates causal confusion, helping the robot to efficiently learn the demonstrated task and successfully execute it even after the beacons are removed.
Future work will explore integration of different beacon types such as auditory and haptic in tasks involving multiple relevant objects and features.

\balance
\bibliography{references} 
\bibliographystyle{ieeetr}

\end{document}